\documentclass[10pt, conference, compsoc]{IEEEtran}

\usepackage[utf8]{inputenc}     
\usepackage[T1]{fontenc}          
\usepackage{amsmath}            
\usepackage{amssymb}             
\usepackage{amsfonts}         
\usepackage{bm}                   
\usepackage{graphicx}             
\usepackage{float}                
\usepackage{stfloats}             
\usepackage{booktabs}             
\usepackage{microtype}            
\usepackage{url}                  
\usepackage{cite}                
\usepackage{lipsum}
\usepackage{amsfonts}
\usepackage{makecell}
\usepackage{pdflscape}
\usepackage{longtable}
\usepackage{booktabs}
\usepackage{hyperref}
\usepackage{titlesec}
\titleformat{\section}[block]{\Large\bfseries\filcenter}{\thesection}{1em}{}

\title{\LARGE \bfseries Constructed Reality, Contested Priors: Decoupling and the Architecture of Cognitive Relapse Under the Free Energy Principle}

\author{
    \IEEEauthorblockN{MD Ibrahim Hossain Ridoy}
    \IEEEauthorblockA{Independent Researcher\\
    Corresponding Author: ibrahimhossainridoy080@gmail.com\\
    ORCID: 0009-0002-7988-9414\\
    (July 2026)}
}

\vspace{10 mm}

\begin{document}

\maketitle
\begin{center}
    \textbf{ABSTRACT}
\end{center}
    
Under the free energy principle, a predictive system does not observe reality directly; it maintains a generative model of the world and experiences that model's best current hypothesis in place of direct sensory contact. This raises a question neuroscience has no way to test directly in a living system: can a synthetic environment be made consistent enough that a predictive system's own inference machinery adopts it as this default hypothesis, permanently displacing the environment that first shaped it? We call this state ontological inversion. Because inducing and monitoring such a transition in a nervous system is neither ethical nor technically feasible, we study the underlying computational problem through a controlled proxy: a convolutional variational autoencoder paired with a recurrent latent predictor, whose evidence lower bound objective is mathematically identical, up to sign, to variational free energy itself. The network is trained first on a baseline visual domain, then on a mixed stream in which a swept rehearsal ratio $r$ controls how much baseline content persists during transition to a target domain. Representational capacity, what the latent space can discriminate, is tracked separately from default behavior, what the system generates when left unconstrained. Across a full sweep of 90 runs, the two diverge sharply: representational accuracy stays near ceiling, 0.97 to 0.998, regardless of $r$, while default behavior spans nearly the system's entire range depending on $r$ alone, a decoupling of learning from acceptance. More strikingly, at intermediate $r$ the system's default output rises toward the target domain, in most runs nearing full capture, then partially reverts toward the baseline while training continues unchanged, a structural failure we term cognitive relapse. Resistance to reality-adoption is not reducible to learning speed; it is a structural property with its own distinct failure modes, established here as a computational existence proof and nothing further.


\vspace{3 mm}
\begin{center}
\section{Introduction}
\label{sec:introduction}
\end{center}
\vspace{0.5 mm}

Neuroscience has converged on a claim that sounds abstract until it is taken literally: the brain does not observe the world, it constructs a generative model of the world and experiences that model's best current hypothesis in place of direct sensory contact \cite{friston2010}. Perception, under this account, is not a window onto reality but a controlled hallucination, held in check by prediction error rather than opened directly onto what is actually out there \cite{seth2021}. This is not a philosophical flourish. It is an operational claim about how cortical inference works, and it carries a structural consequence that is easy to state and uncomfortable to sit with: if what a system experiences as real is a model rather than a direct readout, then in principle that model, and with it what the system treats as real, can be replaced. A generative model is not a fixed window that suddenly shows a different scene. It is closer to a standing hypothesis, continuously defended and continuously revisable, and a standing hypothesis can, under the right pressure, be overturned by a different one.

This paper asks what that structural consequence would actually require, as a literal question rather than a metaphor. Can a synthetic environment be constructed with enough internal consistency that a predictive system's own inference machinery comes to treat it, by default, as the world it expects, fully displacing the environment that trained it in the first place? This is the question Nozick's experience machine posed as a matter of choice, whether a person would elect a machine-generated experience over reality \cite{nozick1974}; the question here is stricter, since it concerns not what a system would choose given the option, but whether its own inference machinery would stop treating that option as live at all. We call this state ontological inversion: the point at which a constructed environment becomes a system's primary generative prior, the reference against which new evidence is evaluated, rather than a hypothesis held provisionally alongside a baseline the system has not yet released. This is a stricter claim than belief revision. A system can come to represent a new environment in exhaustive detail and still hold it at arm's length, evaluating it against a baseline it continues to treat as ground truth. Existing work on rapid belief updating under exposure addresses precisely this weaker case: how a specific belief shifts against new evidence within a model that remains otherwise intact \cite{craske2014}. It says nothing about whether the model itself, the thing doing the evaluating, can be durably replaced. That is the question this paper is built to test.

Testing it directly in a living nervous system is neither ethical nor technically possible: there is no way to install a synthetic generative prior in a brain, monitor its competition against a baseline prior over time, and measure the outcome, without doing exactly the kind of harm the question itself should make anyone hesitant about. What can be tested is the underlying computational problem, stripped of biological implementation, in a system whose internal states are fully observable and whose training history is fully controlled. This is the role played here by a convolutional variational autoencoder paired with a recurrent latent predictor, together referred to as the V+M model \cite{ha2018}. The correspondence is not decorative. A variational autoencoder is trained by minimizing its evidence lower bound, a quantity that is, up to sign, the same mathematical object as the variational free energy invoked above: both trade reconstruction accuracy against the complexity of an approximate posterior to bound surprisal \cite{kingma2013}. Training this network by gradient descent is, in this narrow and literal sense, minimizing free energy directly, not simulating the idea of doing so. The network's self-generated, free-running output, what it produces with no external input constraining it, stands in for a system's default state, modeled on the role attributed to self-referential and internally generated cognition in the biological default mode network \cite{raichle2001,buckner2008,andrewshanna2014}, without claiming that this network's rollout behavior is a model of that network's activity. What follows is a mathematical proxy for the computational problem of reality-adoption, not a biological one, and every claim in this paper is scoped accordingly.

Two findings from this proxy motivate everything that follows, stated here as results rather than assumptions built into the model beforehand. The first is a decoupling of learning from acceptance: a system's capacity to represent the distinction between a baseline and a target environment can sit near ceiling almost immediately, largely as a function of how separable the two environments are, while its default, unconstrained behavior remains governed by something else entirely. The second is cognitive relapse: a system's default output can shift toward a target environment, in most cases coming close to full capture, and then partially revert toward the baseline while training continues under the exact conditions that produced the initial shift, a signature no account of rigidity as merely slow learning would predict. Together these findings reframe cognitive rigidity as a structural property with at least two independent failure modes, rather than a single rate of update, developed formally in Section~\ref{sec:theoretical-framework} and tested against the full experimental sweep in Sections~\ref{sec:computational-experiment} through~\ref{sec:discussion}. None of what follows demonstrates that a biological nervous system behaves this way. It demonstrates that a mechanism derived from first principles, tested against a network whose behavior was not assumed in advance and could have failed to show either pattern, produces results consistent with what that mechanism predicts. That is the scope of the claim, and Section~\ref{sec:disc-boundary} states in full where it stops.

\vspace{3 mm}
\begin{center}
\section{Theoretical Framework}
\label{sec:theoretical-framework}
\end{center}

Under the free energy principle, a system maintains an internal generative model $P(o,s)$ specifying the joint probability of sensory observations $o$ and their hidden causes $s$ \cite{friston2010}. Because the true posterior $P(s\mid o)$ is generally intractable, the system approximates it with a recognition distribution $Q(s\mid\mu)$, where $\mu$ denotes the internal states encoding the current belief. Variational free energy $F$ is the divergence between this approximation and the true posterior, plus the surprisal of the observations themselves:

\begin{equation}
F = \mathrm{KL}\big[Q(s\mid\mu)\,\|\,P(s\mid o)\big] - \log P(o)
\label{eq:free-energy}
\end{equation}

Since KL divergence is non-negative, $F$ upper-bounds surprisal, so minimizing $F$ simultaneously drives the recognition distribution toward the true posterior and suppresses the surprise generated by incoming evidence \cite{fristonkilner2006,fristonstephan2007}. Two consequences follow. Experience is a function of the generative model, not of raw sensory input; what a system represents at any moment is its current best hypothesis, weighted against evidence by the precision assigned to each. And the model is plastic: sustained prediction error, the persistent signal that the current model no longer fits incoming evidence, drives changes to the model's parameters and, at longer timescales, to whatever substrate implements them \cite{fristonstephan2007,friston2017,frankland2005}.

\vspace{3 mm}
\subsection{Ontological Inversion: Definition and Conditions}
\label{sec:ontological-inversion}

Ontological inversion is the state in which a synthetic environment $I_s$, rather than a baseline environment $I_0$, constitutes a system's primary generative prior: the model against which new evidence is evaluated by default, and the model that dominates self-referential and internally generated processing when the system is not driven by external input. This is not the same as a system that can represent $I_s$ convincingly while still holding $I_0$ in reserve as its default. It is the state in which the system's unconstrained, self-generated output is centered on $I_s$ rather than $I_0$.

Three conditions must hold jointly for ontological inversion to occur:

\begin{enumerate}
\item The system's internal representation must support reliably distinguishing $I_s$ from $I_0$, to a degree that exceeds what either representation could achieve in isolation before exposure to the other.
\item The system's default, self-generated output, produced with no external input constraining it, must be centered on $I_s$ rather than $I_0$.
\item Total variational free energy under an $I_s$-dominant model must have fallen to a level consistent with stable prediction, meaning the system has learned $I_s$ well enough that new evidence within $I_s$ generates low surprisal.
\end{enumerate}

None of these three is sufficient alone, and this is the theoretical claim this paper is built to test rather than assume. High representational accuracy for $I_s$, condition 1, does not entail that the system's default output has shifted, condition 2: a system can learn to tell two environments apart without that discriminative capacity determining what it produces when left unconstrained. Nor does a shift in default output guarantee that it is stable: a system's unconstrained output can move toward $I_s$ and still be governed by a free energy landscape, condition 3, in which $I_0$ remains a live competitor rather than a fully displaced alternative. Sections~\ref{sec:contrast-roles} and~\ref{sec:failure-modes} develop what follows from treating condition 1 and condition 2 as genuinely separable, and from treating condition 3 as something that can be reached and then partially lost, rather than reached once and retained.

\subsection{Residual and Concurrent Contrast}
\label{sec:contrast-roles}

Prediction error requires a discrepancy: a mismatch between what the current model expects and what the system receives. A system trained exclusively on $I_s$, with no reference to $I_0$ anywhere in its training signal, would appear at first to have the most direct route to ontological inversion. Whether this is true depends on a distinction this paper's theoretical account treats as central and that prior framings of contrast-driven consolidation have generally left implicit.

Baseline exposure can drive prediction error in two distinct ways. \emph{Residual contrast} is the structure a system already carries from prior exposure to $I_0$, established before any transition protocol begins and present in the system's parameters regardless of what it receives afterward. A system that has already formed a strong internal model of $I_0$ will generate substantial prediction error the moment it is trained on $I_s$, purely because $I_s$ does not fit the model it already has, with no need for $I_0$ to appear anywhere in the ongoing training stream. \emph{Concurrent contrast} is different in kind: it is baseline signal delivered continuously, alongside $I_s$, throughout the transition protocol itself. Its effect does not depend on what the system's parameters already encode; it depends on $I_0$ continuing to arrive as real, current evidence for as long as the protocol runs.

These two roles make different, testable predictions. If only residual contrast matters, a system with strong prior exposure to $I_0$ should be able to complete an inversion toward $I_s$ even with zero concurrent $I_0$ exposure during the transition itself, since the prediction error needed to drive consolidation is already available from the mismatch between existing structure and new evidence. If concurrent contrast also matters, in its own right and not merely as a proxy for residual structure, continuing to deliver real $I_0$ evidence throughout the protocol should continue to generate prediction error that favors $I_0$ as a candidate model for as long as that evidence keeps arriving, including after the system's default output has already shifted toward $I_s$. Under this second reading, concurrent contrast does not simply slow an inversion in progress. It remains a live competing pressure against whatever has already been consolidated, for the full duration it is present.

The experiment in Section~\ref{sec:computational-experiment} is built to separate these two readings directly, by holding residual contrast constant, every condition begins from the same baseline-trained starting point, while varying concurrent contrast as a continuous, swept parameter rather than a single fixed value assumed in advance to be optimal.

\subsection{Two Predicted Failure Modes: Decoupling and Relapse}
\label{sec:failure-modes}

Two specific predictions follow directly from Sections~\ref{sec:ontological-inversion} and~\ref{sec:contrast-roles}, and both are stated here as claims the experiment is designed to test, not as descriptions of results already known.

\textbf{Decoupling of learning and acceptance.} Because condition 1 and condition 2 in Section~\ref{sec:ontological-inversion} are logically independent, a system's representational capacity should be measurable separately from its default behavior, and the two should not be assumed to track one another. In particular, if $I_0$ and $I_s$ are sufficiently distinguishable as classes of input in the first place, a system may reach near-ceiling representational accuracy almost immediately, largely as a function of how separable the two domains are, well before its default output shows any corresponding shift. Representational accuracy under this account is a necessary condition for inversion in only the trivial sense that a system must be able to represent a distinction before that distinction can inform its default state; it should not be read as evidence that the default state has moved.

\textbf{Cognitive relapse.} If concurrent contrast functions as a continuously applied competing pressure rather than a one-time trigger, as argued in Section~\ref{sec:contrast-roles}, a system's default output should not be expected to shift and then remain fixed once condition 2 is first satisfied. It should instead remain contestable for as long as concurrent $I_0$ exposure continues, with the shifted default vulnerable to measurable regression back toward $I_0$ under sustained contact with it. This predicts a specific, checkable signature in any system where concurrent contrast is varied continuously: conditions with more concurrent contrast should show not only a slower initial shift toward $I_s$, but a larger and more consistent regression away from any shift that is achieved, rather than simply a scaled-down version of the same monotonic climb seen at low concurrent contrast.

\subsection{Cognitive Rigidity Redefined}
\label{sec:rigidity-redefined}

Under the account developed in Sections~\ref{sec:contrast-roles} and~\ref{sec:failure-modes}, cognitive rigidity is not adequately captured by treating a system as simply slow to update its default state, with rigidity and speed sitting on the same continuum and differing only in rate. A system that shows the decoupling predicted above is not slow to accept $I_s$; it is fully capable of representing $I_s$ and still does not adopt it as a default, which is a difference in kind, not degree. A system that shows the relapse predicted above is not slow either; it can reach a state that looks, at some intermediate point, like a completed inversion, and then move away from that state under continued conditions that produced it in the first place.

This paper treats cognitive rigidity as better characterized along two axes rather than one. The first is a ceiling on default-state capture that persists regardless of representational accuracy, the decoupling axis. The second is a boundary in the amount of concurrent contrast a system can be exposed to before an achieved shift becomes unstable rather than merely slower to reach, the relapse axis. A system can be rigid on either axis independently of the other, and a full account of resistance to ontological inversion has to characterize both rather than treating either as a stand-in for the other.

\subsection{Scope of the Computational Analogy: From Free Energy to the Evidence Lower Bound}
\label{sec:computational-analogy}

Every claim in Sections~\ref{sec:ontological-inversion} through~\ref{sec:rigidity-redefined} is stated at the level of a generative system in general, not a biological nervous system specifically. The experiment reported in Section~\ref{sec:computational-experiment} tests these claims using a structural analogy: an artificial neural network trained by gradient descent, standing in for the generative system described above, with $I_0$ instantiated as one real image domain and $I_s$ as a second, and the rehearsal ratio $r$, the fraction of each training batch drawn from $I_0$ during the transition protocol, operationalizing concurrent contrast directly as a continuously swept parameter rather than a value fixed in advance.

The analogy is grounded in a specific mathematical correspondence rather than a loose metaphor, introduced in Section~\ref{sec:introduction}: a variational autoencoder's evidence lower bound is, up to sign, the same object as the variational free energy in Equation~\ref{eq:free-energy} \cite{kingma2013}. This correspondence is what licenses treating a trained VAE's ELBO as a direct, if narrow, computational instantiation of Equation~\ref{eq:free-energy}, not merely an unrelated quantity that happens to share the paper's vocabulary.

The analogy does not extend further than this. The network described in Section~\ref{sec:computational-experiment} has no cortical hierarchy, no sleep-dependent consolidation, no homeostatic regulation, and no subjective experience. Its self-generated output, produced by running its own latent predictor with no external input, is used as an operational stand-in for condition 2 in Section~\ref{sec:ontological-inversion}, modeled on the role the default mode network plays in self-referential and internally generated cognition in a biological brain \cite{raichle2001,buckner2008,andrewshanna2014}, not as a claim that the network's rollout behavior is a model of that network's activity. Every quantity reported in Sections~\ref{sec:computational-experiment} through~\ref{sec:discussion} is a measurement taken from this artificial system. None of it is a measurement of, or a claim about, biological tissue.

\subsection{Terminology Reference}
\label{sec:terminology}

Table~\ref{tab:terminology} maps the coined terms used in this paper to their nearest existing concept, to keep clear what is standard and what this paper adds.

\begin{table}[H]
\centering
\caption{Terminology used in this paper mapped to standard concepts.}
\label{tab:terminology}
\begin{tabular}{@{}p{2.1cm}p{2.6cm}p{2.6cm}@{}}
\toprule
Term & Standard concept & Novel claim, if any \\
\midrule
Ontological inversion & Dominant-prior replacement in active inference & Default-state capture as a distinct, separately testable condition from representational capacity \\
Residual contrast & Prior-shaped model structure & Naming this as one of two distinct roles retained baseline exposure can play \\
Concurrent contrast & Ongoing prediction-error signal from ground-truth exposure & Naming this as the second, separable role, and predicting it can destabilize an achieved shift \\
Decoupling of learning and acceptance & Representation-behavior dissociation & Framing this as a specific, checkable prediction of the three-condition account, not an incidental finding \\
Cognitive relapse & Regression of an updated model under continued contact with prior evidence & The specific prediction that this regression scales with the amount of concurrent contrast retained \\
$I_0$ / $I_s$ & Baseline environment / synthetic target environment & Standard placeholder notation; no mechanism claim \\
\bottomrule
\end{tabular}
\end{table}

\vspace{3 mm}
\begin{center}
\section{Computational Experiment}
\label{sec:computational-experiment}
\end{center}

The V+M model introduced in Section~\ref{sec:introduction} is specified here in full. All quantities reported below, including generative quality, representational separability, and default-mode behavior, are computed from the trained weights of this network under gradient descent. Nothing about convergence, threshold-crossing, or the shape of any reported curve is guaranteed by the functional form of an equation; it is measured from a model that could, in principle, have failed to show any of the reported patterns.

\subsection{Architecture}
\label{sec:v3-architecture}

The V component is a convolutional variational autoencoder operating on $28\times28$ grayscale images. The encoder applies two stride-2 convolutions (1 to 32 channels, then 32 to 64 channels, kernel size 3, padding 1), flattens the resulting $64\times7\times7$ feature map to a 3{,}136-dimensional vector, and projects it through two parallel linear layers to a mean $\mu$ and log-variance $\log\sigma^2$, each of dimension 32. The decoder mirrors this in reverse: a linear layer expands the 32-dimensional latent sample back to 3{,}136 dimensions, reshaped to $64\times7\times7$, followed by two transposed convolutions (64 to 32, then 32 to 1 channel) restoring the original $28\times28$ resolution, with a sigmoid output. The full V component has 341{,}825 trainable parameters.

The M component is a single-layer GRU with input and output dimension 32 (matching the VAE's latent dimension) and hidden size 256, followed by a linear projection from the 256-dimensional hidden state back to 32 dimensions. This gives the M component 230{,}944 trainable parameters. Combined, V and M total 572{,}769 trainable parameters, deliberately kept small enough to train the full sweep described in Section~\ref{sec:v3-sweep} on a single consumer-grade GPU.

Reconstruction loss is binary cross-entropy, summed over pixels and averaged over the batch. This is chosen over a Gaussian (MSE) decoder because MNIST and FashionMNIST pixel intensities concentrate near 0 and 1, and the implicit Bernoulli likelihood in BCE is a closer probabilistic fit to near-binary pixel data \cite{kingma2013}. The KL term is the closed-form divergence of the diagonal Gaussian posterior $q(z\mid x)=\mathcal{N}(\mu,\mathrm{diag}(\sigma^2))$ against a standard normal prior, with weight $\beta=1$ (no annealing schedule, no $\beta$-VAE scaling beyond unity).

The GRU is trained on pseudo-sequences constructed from the VAE's encoder means. Within each training batch, the $\mu$ vectors are detached from the VAE's computational graph, so the GRU's prediction loss cannot alter the encoder's gradients, and reshaped into groups of 8 consecutive vectors. Each group of 8 yields 7 next-step prediction targets, with the GRU trained via teacher forcing to minimize the mean squared error between its predicted and actual next-step latent vectors. The two networks share a single Adam optimizer (learning rate $10^{-3}$, batch size 128) and are trained jointly; the total loss is the unweighted sum of the VAE loss and the GRU loss. Because the GRU's gradient never reaches the VAE encoder, the representation-learning objective and the sequence-prediction objective are optimized jointly in wall-clock time but are not coupled through backpropagation.

\subsection{Environments and Data}
\label{sec:v3-environments}

Environment A (baseline) is MNIST; Environment B (target) is FashionMNIST. Both are $28\times28$ grayscale image sets with pixel values scaled to $[0,1]$ and no further preprocessing or augmentation, so that neither domain is given an advantage by differential preprocessing. Both datasets use their standard 60{,}000-image training partition and 10{,}000-image test partition; the training partition is further split into the first 50{,}000 images for training and the last 10{,}000 for validation. MNIST and FashionMNIST were chosen for tractability, not because they resemble any realistic pair of competing environments; this choice is revisited in Section~\ref{sec:limitations}.

\subsection{Training Protocol}
\label{sec:v3-training}

Training proceeds in two stages.

\textbf{Stage 1 (baseline).} For each of 15 random seeds, a fresh V+M model is trained on Environment A only, with the Adam optimizer described above, for up to 50 epochs with early stopping (patience of 5 epochs without improvement in combined validation loss). The checkpoint at the best validation epoch is saved and reused as the starting point for every Stage 2 condition run under that seed.

\textbf{Stage 2 (rehearsal).} Training continues from the Stage 1 checkpoint on a mixed data stream. Each training batch of 128 images draws a fraction $r$ from Environment A and $(1-r)$ from Environment B, combined and shuffled within the batch, following experience-replay methodology \cite{rolnick2019}. Stage 2 runs for a fixed 15 epochs (indexed 0 through 14) with no early stopping, and the optimizer state is reinitialized rather than carried over from Stage 1, so that single-domain momentum from the baseline phase does not bias the mixed-domain phase. Validation loss during Stage 2 is computed on a held-out stream mixed at the same ratio $r$ as training.

All random seeds (Python, NumPy, PyTorch, and CUDA where applicable) are set from a single seed value per run, with cuDNN's deterministic mode enabled and its benchmark mode disabled, trading some training speed for exact reproducibility. Seed is therefore the only source of stochastic variation between two runs sharing the same $r$.

\subsection{Metrics}
\label{sec:v3-metrics}

Three quantities are recorded at every epoch of Stage 2.

\textbf{F\_proxy.} The VAE-only ELBO (reconstruction plus KL, excluding the GRU term) evaluated on the $r$-matched validation stream. Because this validation stream's domain composition tracks $r$, and FashionMNIST reconstructs to a substantially higher BCE than MNIST at comparable training progress, differences in F\_proxy across $r$ partly reflect the changing difficulty of the validation mixture itself rather than a domain-independent measure of generative quality. F\_proxy should be read as generative fit to the current training mixture, not as a fixed-content quality score.

\textbf{w\_proxy.} The 5-fold stratified cross-validated accuracy of a logistic regression classifier (scikit-learn, lbfgs solver), retrained from scratch at every epoch on the VAE encoder's $\mu$ vectors for the full MNIST and FashionMNIST test sets (10{,}000 images each, never mixed or subset by $r$), labeled by domain of origin. This follows the linear classifier probe methodology of Alain and Bengio \cite{alain2016} and measures how linearly separable the two domains are in the current latent space, independent of the training mixture.

\textbf{dmn\_proxy.} The mean $P(\text{environment}=B)$ assigned by that same epoch's fitted probe to images generated by the GRU running freely, with no real sensory input after initialization. Ten independent 50-step rollouts are generated per epoch, each seeded from an independently drawn starting latent vector; at every step the GRU's own prediction becomes the next input, the resulting latent is decoded to image space by the VAE decoder, then re-encoded to obtain the $\mu$ vector the probe classifies. The reported value is the mean $P(B)$ pooled across all 500 generated frames (10 rollouts $\times$ 50 steps) for that epoch. dmn\_proxy is the paper's operational measure of default-mode behavior: what domain the network's unconditioned, self-generated output resembles, as distinct from what it is capable of discriminating given real input.

\subsection{Experimental Sweep}
\label{sec:v3-sweep}

The rehearsal ratio $r$ is swept over $\{0.0, 0.20, 0.30, 0.40, 0.50, 1.0\}$, crossed with 15 seeds, giving 90 Stage 2 runs, plus 15 Stage 1 baseline runs, one per seed, reused across all six $r$ conditions for that seed. At 15 epochs per Stage 2 run this yields 1{,}350 per-epoch records across the three tracked metrics. $r=0.0$ is a negative control in one direction: an immediate, unmixed switch to Environment B with zero rehearsal. $r=1.0$ is a negative control in the other direction: continued training on Environment A only, with the network never receiving a single Environment B image during Stage 2.

\subsection{Threshold-Crossing Analysis}
\label{sec:v3-threshold}

For each (seed, $r$) run, the first Stage 2 epoch at which dmn\_proxy $\geq 0.90$ is recorded. Runs that never reach 0.90 within the 15-epoch window are retained as right-censored observations rather than dropped: dropping them would remove exactly the slowest-shifting runs from the average and bias the reported mean time-to-crossing downward. Mean and standard deviation of epoch-to-crossing are computed only over runs that crossed, and are always reported together with the fraction of runs that did not, so that no single summary statistic obscures how much of a condition's spread falls outside the observation window. The 0.90 threshold is a fixed analysis-stage convention applied identically across every condition in the sweep.

\vspace{3 mm}
\begin{center}
\section{Results}
\label{sec:Results}
\end{center}

\begin{figure*}[t]
    \centering
    \includegraphics[width=\textwidth]{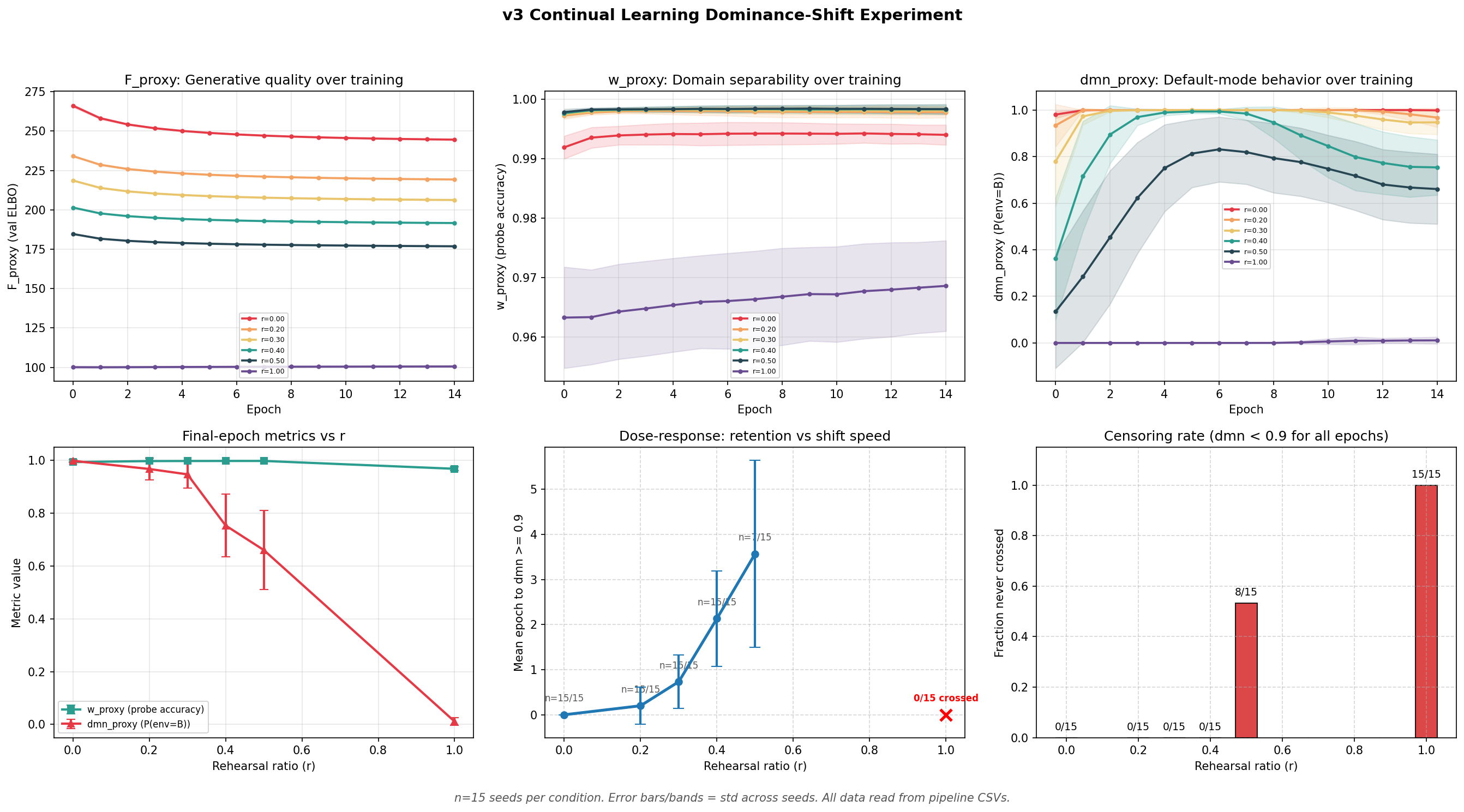}
    \caption{Full sweep across all six rehearsal ratios $r$ and 15 seeds. Top row: F\_proxy, w\_proxy, and dmn\_proxy against Stage 2 epoch, one line per $r$, shaded band is $\pm 1$ standard deviation across seeds. Bottom row: final-epoch (epoch 14) w\_proxy and dmn\_proxy against $r$; mean epoch to dmn\_proxy $\geq 0.90$ against $r$, each point annotated by the fraction of seeds that crossed; fraction of seeds that never crossed 0.90 within the 15-epoch window, against $r$.}
    \label{fig:v3-sweep}
\end{figure*}

All 90 Stage 2 runs completed; no run is excluded from any analysis below on any grounds, including apparent outlier behavior. Figure~\ref{fig:v3-sweep} summarizes the full sweep.

\vspace{3 mm}

\subsection{Onset Speed: Dose-Response in Time to Crossing}
\label{sec:results-onset}

\begin{table}[t]
\centering
\caption{Mean epoch of first dmn\_proxy $\geq 0.90$ crossing, by rehearsal ratio. Mean and standard deviation computed only over seeds that crossed within the 15-epoch window.}
\label{tab:onset}
\begin{tabular}{@{}lccc@{}}
\toprule
$r$ & crossed / total & mean epoch & std \\
\midrule
0.0 & 15/15 & 0.00 & 0.00 \\
0.2 & 15/15 & 0.20 & 0.41 \\
0.3 & 15/15 & 0.73 & 0.59 \\
0.4 & 15/15 & 2.13 & 1.06 \\
0.5 & 7/15  & 3.57 & 2.07 \\
1.0 & 0/15  & --   & --   \\
\bottomrule
\end{tabular}
\end{table}

At $r=0.0$, dmn\_proxy is already at or above 0.90 in every seed at the first evaluated epoch (mean crossing epoch 0.00). Increasing $r$ delays this monotonically across every value tested through $r=0.4$: mean crossing epoch rises to 0.20 at $r=0.2$, 0.73 at $r=0.3$, and 2.13 at $r=0.4$, with all 15 seeds crossing at every one of these four ratios. This is a genuine, monotonic dose-response in onset speed, a property that emerged from gradient descent on a trained network, not a property built into an equation's functional form.

\subsection{Trajectory Shape: Overshoot and Partial Regression}
\label{sec:results-overshoot}

The onset-speed statistic in Table~\ref{tab:onset} describes only the first time dmn\_proxy crosses 0.90; it does not describe what the trajectory does afterward. Examined epoch by epoch, dmn\_proxy is not monotonically increasing for $r > 0$. In every condition with $r \in \{0.2, 0.3, 0.4, 0.5\}$, per-seed trajectories rise rapidly over the first several epochs, in most seeds reaching or closely approaching the ceiling of 1.0, and then decline over the remainder of the 15-epoch window.

\begin{table}[t]
\centering
\caption{Peak versus final (epoch 14) dmn\_proxy, by rehearsal ratio, averaged over 15 seeds.}
\label{tab:overshoot}
\begin{tabular}{@{}lcccc@{}}
\toprule
$r$ & mean peak & epoch of peak & mean final & mean decline \\
\midrule
0.0 & 1.0000 & 6.4  & 0.9989 & 0.0011 \\
0.2 & 1.0000 & 7.3  & 0.9679 & 0.0321 \\
0.3 & 1.0000 & 7.9  & 0.9475 & 0.0525 \\
0.4 & 0.9951 & 5.5  & 0.7538 & 0.2413 \\
0.5 & 0.8728 & 6.9  & 0.6608 & 0.2120 \\
1.0 & 0.0152 & 12.3 & 0.0110 & 0.0042 \\
\bottomrule
\end{tabular}
\end{table}

The size of this decline scales with $r$. It is small at $r=0.2$ (mean drop 0.032 from peak to epoch 14) and $r=0.3$ (0.053), then increases sharply at $r=0.4$ (0.241, with 14 of 15 seeds declining more than 0.03 from their individual peak, one by as much as 0.42) and remains comparably large at $r=0.5$ (0.212). $r=1.0$ shows no equivalent structure: its nominal peak (mean 0.015, at a mean epoch of 12.3) sits inside the noise floor the metric holds throughout Stage 2, not a genuine excursion.

This means $r=0.4$ and $r=0.5$ are not simply slower versions of the pattern seen at $r=0.0$ through $r=0.3$. In both conditions, the network's free-running output shifts toward Environment B early in Stage 2 and then partially reverts toward Environment A dominance while training continues, with the size of the reversion increasing as more Environment A content is retained in the training stream. The 15-epoch window used throughout this sweep is not long enough to determine whether this regression stabilizes at some intermediate value, continues indefinitely, or eventually reverses again; that question is outside the scope of the runs reported here.

\subsection{The $r=0.5$ Condition}
\label{sec:results-r05}

Seven of 15 seeds cross the 0.90 threshold (mean epoch 3.57, std 2.07). The remaining eight do not cross within the 15-epoch window; their epoch-14 dmn\_proxy values range from 0.361 to 0.895 (mean 0.611). Inspecting the full trajectories of these eight seeds, rather than only their final value, shows that seven of the eight have already passed a higher peak earlier in Stage 2 (individual peaks ranging from 0.673 to 0.899, all below 0.90, with two peaks within 0.01 of it and the rest between 0.04 and 0.23 below it) and are flat or declining by epoch 14. Only one of the eight (final value 0.895, still rising at epoch 14) shows a trajectory consistent with continuing to climb toward threshold. The remaining seven do not support the inference that additional epochs would produce a crossing; several show the same peak-then-regress pattern documented in Section~\ref{sec:results-overshoot}, and their epoch-14 trajectories are flat or falling, not rising.

\subsection{The $r=1.0$ Condition: Complete Block}
\label{sec:results-r10}

Zero of 15 seeds cross the 0.90 threshold. Mean dmn\_proxy at epoch 14 is 0.0110 (std 0.0146), close to its starting value and essentially unmoved across all 15 epochs of Stage 2. This is the expected result for a condition in which the network receives zero Environment B images throughout Stage 2, and it held without exception.

Notably, w\_proxy at $r=1.0$ still reaches 0.9686 (std 0.0076) by epoch 14, the lowest w\_proxy value anywhere in the sweep but far above chance (0.5), despite the VAE encoder never seeing a single FashionMNIST image during Stage 2. This indicates that MNIST and FashionMNIST are separable enough as image classes that a linear probe can distinguish them in this network's latent space even when the encoder was never trained to make that particular distinction; see Section~\ref{sec:results-decoupling}.

\subsection{Decoupling of Representational Separability from Default-Mode Behavior}
\label{sec:results-decoupling}

\begin{table}[t]
\centering
\caption{Final-epoch (epoch 14) w\_proxy and dmn\_proxy, mean $\pm$ std across 15 seeds, by rehearsal ratio.}
\label{tab:decoupling}
\begin{tabular}{@{}lcc@{}}
\toprule
$r$ & w\_proxy & dmn\_proxy \\
\midrule
0.0 & $0.9940 \pm 0.0017$ & $0.9989 \pm 0.0041$ \\
0.2 & $0.9978 \pm 0.0009$ & $0.9679 \pm 0.0409$ \\
0.3 & $0.9981 \pm 0.0007$ & $0.9475 \pm 0.0522$ \\
0.4 & $0.9983 \pm 0.0008$ & $0.7538 \pm 0.1182$ \\
0.5 & $0.9983 \pm 0.0008$ & $0.6608 \pm 0.1499$ \\
1.0 & $0.9686 \pm 0.0076$ & $0.0110 \pm 0.0146$ \\
\bottomrule
\end{tabular}
\end{table}

Across the entire sweep, w\_proxy ranges from 0.9686 to 0.9983, a span of 0.0297. Over the same sweep, dmn\_proxy ranges from 0.0110 to 0.9989, a span of 0.9879, nearly the full width of the metric's possible range. Representational separability in the VAE's latent space is close to ceiling in every tested condition, including $r=1.0$, where the model receives no real Environment B input at all during Stage 2. Whether the network's free-running, self-generated output resembles Environment A or Environment B is answered almost independently of whether its latent space can, given real input, tell the two domains apart. Two conditions can share a w\_proxy within 0.0002 of each other ($r=0.4$ at 0.9983 and $r=0.5$ at 0.9983) while differing in dmn\_proxy by nearly 0.1 (0.7538 versus 0.6608), and $r=1.0$, with the single lowest w\_proxy in the sweep, has a dmn\_proxy roughly two orders of magnitude smaller than every other condition. High w\_proxy is necessary in the trivial sense that the probe needs a real signal to classify against; it is nowhere close to sufficient to predict dmn\_proxy.

\vspace{3 mm}
\begin{center}
\section{Discussion}
\label{sec:discussion}
\end{center}

The central question of this paper, stated in Section~\ref{sec:ontological-inversion}, is whether a synthetic environment can be engineered to displace a baseline one as a generative system's default reference, not just as something the system can represent when asked, but as what its own unconstrained output defaults to. The V+M results in Section~\ref{sec:Results} do not answer that question for a biological system. They answer a narrower, computational version of it, and the answer is more structured than the sweep's headline dose-response result suggests: the shift can be induced, its onset speed is under direct experimental control, and it is not stable once induced. Each of these three parts carries a different implication, taken up in turn below.

\vspace{3 mm}
\subsection{Cognitive Relapse and the $r=0.5$ Transition Regime}
\label{sec:disc-relapse}

A pure slowdown account of the sweep would predict that dmn\_proxy at every tested $r$ rises monotonically toward a plateau, with the rate of rise set by $r$ and nothing else changing in shape. Section~\ref{sec:results-overshoot} shows this is not what happens. At $r \in \{0.2, 0.3, 0.4, 0.5\}$, dmn\_proxy overshoots, in most seeds reaching or approaching 1.0, and then declines for the remainder of the 15-epoch window, with the size of the decline scaling from negligible at $r=0.2$ to substantial at $r=0.4$ and $r=0.5$ (Table~\ref{tab:overshoot}). This is not slow learning. Slow learning would look like a shallower version of the same curve. What is observed instead is a system that reaches a state resembling adoption of Environment B and then measurably moves away from it while training continues under the same regime that produced the adoption in the first place. This transient regression resembles the "stability gap" documented in continual learning \cite{delange2023}, though here it manifests in unconditioned generative output rather than supervised task accuracy.

The mechanism is not mysterious given the training protocol. At every $r>0$, a fraction $r$ of every batch in every epoch of Stage 2, not only the first few epochs, is real Environment A data, updating the same shared weights that produce the free-running rollout. Rehearsal is not a one-time perturbation the system absorbs and moves past. It is a continuous source of gradient pressure toward Environment A for the entire duration of training. Early in Stage 2, the network's unconditioned output can apparently drift toward Environment B faster than the accumulated pull of the retained Environment A batches, only for that pull to reassert itself as training continues. We refer to this descriptively as \emph{cognitive relapse}: an adopted default state that is not maintained against continued contact with the source it displaced. The term labels a specific curve shape, rise, peak, decline, measured in an artificial network. It is not a claim about anything psychological; Section~\ref{sec:disc-boundary} states this restriction explicitly.

The $r=0.5$ condition marks a qualitative boundary within the tested grid, not only a point further along a smooth trend. It is the first ratio at which a majority of seeds, 8 of 15, fail to reach the 0.90 threshold at all within the window, and cross-seed variance rises sharply alongside it: standard deviation of epoch-to-crossing goes from 1.06 at $r=0.4$ to 2.07 at $r=0.5$, and standard deviation of final-epoch dmn\_proxy goes from 0.118 to 0.150 over the same step. Rising variance near a boundary is consistent with, though it does not on its own establish, an approach to a genuine bifurcation in the underlying dynamics, where nearby initial conditions or noise realizations begin to produce qualitatively different long-run outcomes. Whether $r=0.5$ sits on a true phase transition, or is simply the last point before a smooth continued decline into the untested interval between $r=0.5$ and $r=1.0$, cannot be resolved by this sweep: no runs exist between those two values. Confirming a genuine transition would require sampling that interval directly and checking whether the fraction of seeds reaching threshold falls off sharply over a narrow sub-range of $r$ or declines gradually across it.

\subsection{The Decoupling of Learning and Acceptance}
\label{sec:disc-decoupling}

Section~\ref{sec:results-decoupling} shows w\_proxy confined to a narrow band, 0.9686 to 0.9983, across every tested condition, while dmn\_proxy spans nearly the metric's full range, 0.0110 to 0.9989, over the same sweep. This is the clearest empirical instance in this paper of a claim that Section~\ref{sec:ontological-inversion} made on theoretical grounds alone: that the conditions for a generative prior to shift are not reducible to one another. A network's capacity to represent the distinction between two environments and the environment its own unconstrained behavior defaults to are, in this system, close to independent quantities. At $r=1.0$, the network never receives a single Environment B image during Stage 2, yet w\_proxy still reaches 0.9686 there, meaning its latent space can still separate the two domains at high accuracy under real input, while dmn\_proxy sits at 0.0110, meaning its unconditioned output almost never resembles Environment B. Representational capacity is not a precursor state that default behavior eventually catches up to in this data. It is close to saturated everywhere, including where default behavior does not move at all.

This distinction maps onto a difference the Introduction already draws between believing a constructed environment is real while still holding it at arm's length, and the constructed environment being what the system expects by default, with no reference held in reserve~\cite{seth2021}. w\_proxy indexes something closer to the first sense: whether the discriminative machinery exists to tell the two apart. dmn\_proxy indexes something closer to the second: what the system's generative process treats as its baseline output when nothing external constrains it. These results show that the two can be pulled almost fully apart. Training a system to represent an alternative environment well is not evidence that the system's default behavior has moved toward it, and the two should not be conflated when either is reported as evidence for anything resembling reality displacement.

\subsection{What This Does and Does Not Say About Ontological Inversion}
\label{sec:disc-implications}

Taken together, Sections~\ref{sec:disc-relapse} and~\ref{sec:disc-decoupling} narrow what a computational demonstration of ontological inversion would need to show. It is not enough to demonstrate that a system can be trained to represent a synthetic environment; w\_proxy is high everywhere in this sweep, including the negative control, and represents nothing about default behavior on its own. It is not enough to demonstrate an initial shift in default behavior, either: two of the six tested conditions in this sweep show an initial shift that substantially reverses within the same 15-epoch window. Under this stricter reading, only $r=0.0$ through $r=0.3$ show something resembling a stable capture of default behavior within the tested window. $r=0.4$ and $r=0.5$ show partial capture followed by relapse. $r=1.0$ shows no capture at all. None of the six conditions was run past epoch 14, so whether the $r=0.4$ and $r=0.5$ relapse trajectories eventually re-stabilize at some intermediate value, continue reverting toward the $r=1.0$ floor, or oscillate is not something this experiment can answer.

This bears on the contrast-retention argument raised earlier in the paper, that some ongoing exposure to the baseline environment is necessary to sustain the prediction-error gradient that drives consolidation of a new one. The rehearsal ratio $r$ in this experiment operationalizes that same general idea, more baseline retention should mean more sustained contrast, in a different architecture and without asserting the specific mechanism proposed for it there. These results complicate a simple reading of that argument rather than confirming it outright. Retaining more baseline content does not only slow the shift toward the new environment, which a contrast-as-fuel framing predicts; it also appears to actively contest an already-partially-completed shift throughout training, which a pure fuel-for-consolidation framing does not predict on its own. Some baseline retention may be necessary to drive an initial shift at all, consistent with $r=0.0$ never showing a moment's hesitation. Too much retention appears able to prevent that shift from holding once achieved. The right amount, and whether a fixed ratio held constant for an entire run is even the correct way to administer it, is not something this sweep was designed to
determine. This dynamic is structurally adjacent to "wireheading" in reinforcement learning, where agents exploit self-generated signals over real environments \cite{ringorseau2011}. However, while wireheading requires an agent to actively choose delusion for reward optimization, ontological inversion here occurs as a passive transition in the model's default generative state under externally imposed exposure.

\subsection{Scope of the Structural Analogy}
\label{sec:disc-boundary}

Everything reported in this Discussion describes a 572{,}769-parameter VAE and GRU trained on MNIST and FashionMNIST. It has no analog of a cortical hierarchy, no sleep-dependent consolidation, no homeostatic regulation, no competing top-down attentional control, and no subjective experience of any kind. dmn\_proxy measures the classification of pixels produced by an unconditioned rollout of a next-latent predictor; it is not a measurement of activity in a biological default mode network. w\_proxy measures cross-validated logistic regression accuracy, not synaptic weight. "Relapse," "acceptance," and "default behavior," as used above, are labels for the shape of a measured curve in this specific system, chosen because they are the most direct available language for that shape, not because the underlying process is claimed to resemble relapse, acceptance, or defaulting as those terms are used for a nervous system.

What this experiment supports is narrower and more specific. Two claims made independently of any dynamical-systems construction in this paper's Theoretical Framework, that representational capacity and default generative behavior can dissociate, and that a system moved toward a new default under sustained partial exposure to its prior one does not necessarily hold that default once achieved, both survive contact with a trained, gradient-descent-optimized network in which neither property was assumed or built in by hand. That is evidence the theoretical account in Section~\ref{sec:ontological-inversion} is not merely restating itself in different words when tested against a second, independently constructed system. It is not evidence about what a biological brain does under any comparable protocol, and no claim to that effect is made here or should be drawn from it.

\vspace{3 mm}
\begin{center}
\section{Limitations}
\label{sec:limitations}
\end{center}

This section consolidates every scope and limitation statement made in this paper into one place, so that what this work does not support is available from a single location rather than scattered across sections as qualifications easy to skip past.

\subsection{The Fifteen-Epoch Window Is Fixed and Untested Beyond That Point}
\label{sec:lim-window}

The cognitive relapse pattern documented in Section~\ref{sec:results-overshoot} and interpreted in Section~\ref{sec:disc-relapse} is observed only within a fixed 15-epoch Stage 2 window. At $r=0.4$ and $r=0.5$, dmn\_proxy rises, peaks, and declines within that window; what happens beyond epoch 14 is not tested. Three distinct continuations are consistent with the recorded data and cannot be distinguished by this experiment. The decline could stabilize at some intermediate value, held by a balance between concurrent contrast pulling toward $I_0$ and whatever remains of the pull toward $I_s$. It could continue declining toward the $r=1.0$ floor, meaning the observed shift was transient and sustained rehearsal eventually erases it entirely. Or it could reverse again, since nothing in the training protocol changes after epoch 14 and the same forces that produced the initial rise are still present in every batch that follows. Resolving which of these is correct requires running the $r=0.4$ and $r=0.5$ conditions substantially longer, at minimum long enough to observe whether the decline rate is itself decaying, consistent with stabilization, or holding constant or accelerating, consistent with continued or complete reversal. No run in this sweep was extended far enough to make that determination. "Relapse," as used in Section~\ref{sec:disc-relapse}, describes the shape observed within the tested window. It is not a claim about that shape's long-run destination.

\subsection{MNIST and FashionMNIST Are Not a Realistic Pair of Competing Environments}
\label{sec:lim-domains}

MNIST and FashionMNIST were chosen for computational tractability and fast iteration, not because they resemble any realistic pair of environments a generative system might be asked to choose between. The two domains are separable almost trivially: w\_proxy exceeds 0.96 in every condition tested, including $r=1.0$, where the encoder never receives a single FashionMNIST image during Stage 2. This bears directly on how the decoupling result in Sections~\ref{sec:results-decoupling} and~\ref{sec:disc-decoupling} should be read. The near-ceiling, nearly $r$-invariant w\_proxy observed across this sweep may be substantially a property of how easy this particular pair of domains is to separate, rather than a general property of representational learning under domain transition. A pair of environments with genuinely overlapping low-level statistics, more visually or structurally similar to one another than a digit is to a garment, might show w\_proxy itself varying meaningfully with $r$, in which case the dissociation between w\_proxy and dmn\_proxy reported here could narrow. That learning and acceptance can decouple is supported by this experiment. That they decouple this sharply, or under any sufficiently different pair of environments, is not.

\subsection{The Rehearsal Ratio Is a Crude Operationalization of Concurrent Contrast}
\label{sec:lim-r}

The rehearsal ratio $r$, as implemented in Section~\ref{sec:v3-training}, is a single scalar controlling the fraction of each training batch drawn from Environment A, held fixed for the entire 15-epoch Stage 2 window and applied identically to every batch. This is a coarse instantiation of concurrent contrast as defined in Section~\ref{sec:contrast-roles}. It does not vary the timing, patterning, or salience of retained baseline exposure within a run, whether it arrives early or late, clustered or interspersed, attended to or incidental, all of which plausibly matter for anything resembling a real contrast signal and are not represented by a ratio held constant throughout training. It also treats every unit of Environment A content as interchangeable, with no analog of selective attention or relevance determining how much prediction-error signal any given piece of retained content actually generates. $r$ should be read as the simplest operationalization of concurrent contrast sufficient to test whether the concept does anything at all, not as a considered model of how contrast operates. That it produces the graded onset delay and the $r$-dependent relapse magnitude reported in Section~\ref{sec:Results} despite this simplicity is informative. It is not evidence that batch-fraction mixing is the correct or only way concurrent contrast could be built into a system.

\subsection{Residual Contrast Was Held Constant, Not Varied}
\label{sec:lim-residual}

Every condition in this sweep begins from the same Stage 1 procedure, up to 50 epochs on Environment A alone with early stopping at patience 5, identical across every value of $r$ for a given seed. This holds residual contrast, as defined in Section~\ref{sec:contrast-roles}, constant across the entire sweep, which is what allows the sweep to isolate the effect of concurrent contrast cleanly. It also means residual contrast's own contribution was never independently tested. Whether a system with a weaker Stage 1, less thoroughly consolidated on Environment A to begin with, would show a different pattern of onset speed or relapse magnitude under the same $r$ sweep is not addressed here. The distinction between residual and concurrent contrast is supported in this experiment only in the sense that concurrent contrast alone, with residual contrast fixed, produces the reported effects. It has not been shown that residual contrast strength is itself a free variable with its own independent effect on either onset speed or relapse.

\subsection{Single Architecture and Fixed Analysis Thresholds}
\label{sec:lim-architecture}

Latent dimension, GRU hidden size, learning rate, batch size, and sequence length were fixed before the sweep and not varied; whether the reported patterns hold under different architectural choices is untested. The 0.90 threshold used for the epoch-to-crossing analysis in Section~\ref{sec:v3-threshold} is a single fixed convention, not swept across alternatives. A lower or higher threshold would shift the specific onset-epoch and censoring numbers reported in Table~\ref{tab:onset}, though it is unlikely to remove the qualitative overshoot-and-decline pattern in Table~\ref{tab:overshoot}, since that pattern is visible in the raw per-epoch trajectories independent of any particular threshold choice.

\subsection{What This Model Does Not Represent}
\label{sec:lim-scope}

No condition beyond $r=0.0$ and $r=1.0$ isolates batch composition from other factors that covary with it, such as total accumulated gradient signal from each domain across the full 15 epochs. Every seed starts from an architecturally identical, randomly initialized network differing only in its random seed; there is no representation of individual variation in prior structure or differential susceptibility to either domain. dmn\_proxy and w\_proxy remain abstract indices bounded in $[0,1]$, calibrated against nothing beyond the classifier and rollout procedure that produce them; a value of 0.75 has meaning only relative to other values in this sweep, not against any external reference point. No claim is made anywhere in this paper that any measured quantity corresponds to subjective experience in any system, biological or artificial. Convergence, capture, or relapse of dmn\_proxy in this network implies nothing about whether any system would report a synthetic environment as more or less real than a baseline one. That inference would require evidence this experiment was not designed to produce and does not claim to have produced.

\vspace{3 mm}
\begin{center}
    \bfseries\large Conclusion
\end{center}
\vspace{0.5 mm}

The central argument of this paper is that if a generative system's experience of its environment is produced by a model rather than read directly from evidence, that model can, in principle, be redirected toward a different environment than the one that first shaped it. Section~\ref{sec:theoretical-framework} restated this as three jointly necessary conditions and derived two specific predictions from them: that representational capacity and default generative behavior should be separable rather than coupled, and that a shift in default behavior driven by concurrent contrast should remain contestable for as long as that contrast continues, rather than locking in permanently the first time it is observed.

Both predictions held. Across the full sweep in Section~\ref{sec:Results}, w\_proxy stayed within a narrow band of 0.9686 to 0.9983 regardless of rehearsal ratio, while dmn\_proxy ranged from 0.0110 to 0.9989 over the same conditions: a system that can represent a target environment almost perfectly while its default output remains governed by something other than that representational capacity. And at rehearsal ratios of 0.4 and 0.5, the network's default output did not simply approach the target environment more slowly than at lower ratios. It rose toward the target, in most seeds coming close to full capture, and then partially reverted toward the baseline while training continued under the same conditions that had produced the initial rise, a pattern with no equivalent at $r=0.0$ or $r=1.0$ and one that a model of rigidity as slow learning alone does not predict.

Neither result was assumed into the model beforehand. dmn\_proxy and w\_proxy are measurements taken from a network trained by ordinary gradient descent on two real image datasets; nothing about their relationship to one another, and nothing about whether a rising trajectory would hold or regress, was fixed by the architecture or the loss function in advance. That both predictions were borne out by a system built without either property assumed is what makes this a computational existence proof rather than a restatement of the theory in different notation. The redefinition of cognitive rigidity proposed in Section~\ref{sec:rigidity-redefined}, as a decoupling axis and a relapse axis rather than a single rate of update, describes something a trained network actually does, not only something a theory says it should do.

This remains, in the strictest sense, exactly that and no more: a computational existence proof. It establishes that decoupling and relapse are not internally inconsistent as concepts, and that a mechanism producing both is not merely conceivable but constructible in a system whose behavior was not guaranteed by design. It does not establish that biological synaptic weighting decouples from default mode network engagement the way w\_proxy decouples from dmn\_proxy here, or that any nervous system exposed to sustained partial contrast would show a comparable rise and partial reversion in whatever its own analog of default behavior would be. Confirming either would require evidence this experiment cannot produce: measured shifts in neural population coding for two competing internal models, and a longitudinal account of how self-generated activity in a real brain responds to sustained, partial exposure to a displaced prior. Section~\ref{sec:limitations} states in full what stands between this result and that evidence. None of it is minimized here.

What this paper adds is narrower and more concrete than a validation would be. It shows that a specific, falsifiable account of resistance to model replacement, built on separating representational capacity from default behavior and on treating retained contrast as a continuous pressure rather than a one-time trigger, can be operationalized in a trained system, tested against a full parameter sweep rather than a single run, and found to hold. That is the claim this paper makes, and it is the only one it makes.

\vspace{3 mm}
\section*{Project Finite Tsukuyomi}
\label{sec:finite-tsukuyomi}

This paper addresses one part of a larger research program, which the author calls Project Finite Tsukuyomi, built on three pillars. Decode is the problem of reading a subject's own preferred reality directly from neural signals, extracting a preference from the activity of the system that holds it rather than inferring it from a menu of predefined choices. Build is the problem of constructing a synthetic environment that matches a decoded preference with enough coherence to sustain the prediction-error minimization this paper's theoretical account depends on, a world that holds together under sustained interaction rather than merely resembling what was decoded. Accept, the subject of this paper, asks what conditions govern whether a generative system durably adopts a given synthetic environment as its default rather than holding it as one hypothesis among several. The vocabulary developed here, residual and concurrent contrast, the decoupling of learning and acceptance, cognitive relapse, is offered as a partial answer to that third question specifically.

Neither Decode nor Build exists as a demonstrated capability today. Neural decoding of preference, as distinct from decoding of intended action or reported affect, has not been shown at the specificity this program requires, and generative world models have not been shown to sustain coherence across the open-ended, multi-session exposure the second pillar would demand. Both fields are moving toward these capabilities, not away from them, but neither has reached the point this program's first two pillars assume as a starting condition. This paper does not depend on either existing yet. It establishes the third pillar's mechanism on its own terms, so that its specific failure modes, decoupling and relapse, are known in advance rather than discovered after the fact in a system where discovery costs more than a discarded training run.

A program built from these three pillars is not made ethically neutral by being incomplete. Decoding a preference a person has not consciously articulated, constructing an environment engineered to match it, and consolidating that environment as a default reality are each, on their own, actions that raise questions of consent, reversibility, and who is authorized to make that decision on whose behalf, questions as fundamental to this program as any question of mechanism. This paper does not resolve them, and a computational analogy run on MNIST and FashionMNIST is not positioned to; they are stated here because a research program is not honestly described by its technical ambition alone. The author welcomes collaboration from anyone working on either remaining pillar, or on the boundary between them, in neural decoding, generative world modeling, or the theoretical and ethical questions the full program raises.

\vspace{2 mm}
\textbf{Acknowledgments :} The author utilized AI assistance (Claude, Anthropic) exclusively for language refinement, formatting, code debugging and structural editing. All experimental design, theoretical frameworks, code execution, data generation, and core scientific conclusions are completely original.

\vspace{3em}
\hrule
\vspace{1.5em}
\textbf{Data and Code Availability :} All code required to reproduce the results reported in this paper is publicly available at [\url{https://github.com/ibrahim-hossain-ridoy/Finite-Tsukuyomi-code-v3}]. This includes the VAE and GRU model definitions, the two-stage training pipeline, the analysis scripts used to compute every summary statistic and threshold-crossing result in Section~\ref{sec:Results}, and the script used to generate Figure~\ref{fig:v3-sweep}. The complete epoch-by-epoch metric record for all 90 Stage 2 runs, 1{,}350 rows spanning seed, rehearsal ratio, epoch, F\_proxy, w\_proxy, and dmn\_proxy, is included in the repository as a single CSV file, alongside the derived summary and threshold-crossing,and relapse-statistics CSVs used to produce every table in this paper. All hyperparameters, random seeds, and the rehearsal ratio grid are fixed in a single configuration file and are not altered between runs. No result reported in this paper depends on data or code withheld from this repository.

\end{document}